\documentclass[letterpaper, 10 pt, conference]{ieeeconf}  

\IEEEoverridecommandlockouts                              

\usepackage{graphics} 
\usepackage{mathptmx} 
\usepackage{amsmath} 
\usepackage{amssymb}  
\usepackage{graphicx}
\usepackage{cite}
\usepackage{multirow}
\usepackage{makecell}
\usepackage[pagebackref,breaklinks,colorlinks]{hyperref}
\usepackage[capitalize]{cleveref}
\usepackage{color}

\title{\LARGE \bf
Channel-Aware Distillation Transformer for Depth Estimation on Nano Drones
}

\author{Ning Zhang$^{*}$, Francesco Nex, George Vosselman, Norman Kerle
\thanks{All the authors are with the ITC Faculty Geo-Information Science and Earth Observation,
        University of Twente, 7514 AE Enschede, The Netherlands.}
\thanks{$^{*}$Correspondence: n.zhang@utwenten.nl
}}

\begin{document}

\maketitle
\thispagestyle{empty}
\pagestyle{empty}

\begin{abstract}

Autonomous navigation of drones using computer vision has achieved promising performance. Nano-sized drones based on edge computing platforms are lightweight, flexible, and cheap, thus suitable for exploring narrow spaces. However, due to their extremely limited computing power and storage, vision algorithms designed for high-performance GPU platforms cannot be used for nano drones. To address this issue this paper presents a lightweight CNN depth estimation network deployed on nano drones for obstacle avoidance. Inspired by Knowledge Distillation (KD), a \textbf{C}hannel-\textbf{A}ware \textbf{Di}stillation \textbf{T}ransformer (CADiT) is proposed to facilitate the small network to learn knowledge from a larger network. The proposed method is validated on the KITTI dataset and tested on a nano drone Crazyflie, with an ultra-low power microprocessor GAP8.

\end{abstract}

\section{INTRODUCTION}
Drones play an important role in exploration tasks. In particular, nano-sized drones are suitable for exploring narrow and cluttered environments after disasters because of their small size and relative affordability~\cite{45,48}. Flying autonomous drones in scenarios where GNSS is not available is a challenging research topic. Some research integrates multiple sensors, such as high-quality stereo cameras~\cite{13,14} and LiDAR~\cite{6,7,8,16}, to navigate based on maps. This type of method requires drones with large payloads~\cite{47}. As computer vision technology evolves, learning-based methods using monocular vision emerge. Given an input image a convolutional neural network (CNN) can be trained to directly output control commands~\cite{1,2,9}. Since this type of method cannot control the drone's next waypoints, it is more conservative to navigate and avoid obstacles based on depth maps predicted by CNNs~\cite{5,4,10}. The advantages of using depth maps for drone navigation are two-fold: i) A depth map intuitively represents the distance from each object to the viewpoint in the scene and is ideal for navigating a drone and selecting the next waypoint if needed. ii) Some recent CNN-based depth estimation methods leverage the self-supervised training strategy and do not require labeled data for training. In view of these advantages the focus of this paper is also on the use of the depth map estimated by a CNN for the obstacle avoidance of nano-drones.

Nevertheless, the target platforms of the aforementioned methods were not nano drones and they used high-performance graphics processing units (GPUs), which are not available in nano drones. For example, Bitcraze's Crazyflie, which uses GAP8 as its processor and is the platform used in this paper, has only 22.65 GOPS of processing power and 512 KB of RAM. This is still not enough for the storage and inference of small models such as Lite-Mono~\cite{11}, while smaller models have limited learning capacity. To address this problem the framework DDND (Distilled Depth for Nano Drones) is proposed, which takes advantage of self-supervised training and knowledge distillation. This paper uses another larger trained model as the teacher model and distills its knowledge into the smaller model. The contribution of this paper can be summed up as follows.
\begin{itemize}
    \item A lightweight depth estimation framework DDND is proposed that has few parameters ($310K$). Knowledge distillation is introduced to improve the learning ability of such a small model, and puts forward the \textbf{C}hannel-\textbf{A}ware \textbf{Di}stillation \textbf{T}ransformer (CADiT) to make the student model explore geometric cues in different feature channels from the teacher, thus enhancing the knowledge distillation. The effectiveness of the method is validated on the KITTI dataset.

    \item The proposed model is deployed on the nano drone platform Crazyflie, and it runs at 1.24 FPS on a GAP8 processor to avoid obstacles in real environments. To our knowledge, this paper is the first to introduce distilled depth estimation for nano drones. The code will be released on the project  \href{https://github.com/noahzn/DDND}{website}.
\end{itemize}

The rest of the paper is organized as follows. Section~\ref{literaturereview} reviews some related research work. Section~\ref{method} describes
the proposed method in detail. Section~\ref{sec:exp} elaborates on the experiments and discussions. Section~\ref{sec:conclusion} concludes the paper.

\section{RELATED WORK}
\label{literaturereview}

\subsection{Obstacle Avoidance of Nano Drones}
Nano drones can be equipped with small laser rangers~\cite{49,50,45,48,56} or sonars~\cite{51} to avoid obstacles at short distances. Some research focused on optical flow estimation using a dedicated optical flow sensor ~\cite{52} or cameras~\cite{53,54}. With the development of edge computing devices, obstacle avoidance using CNNs methods is beginning to make its mark on edge computing platforms like PULP~\cite{55} and Crazyflie. Image-based methods can provide rich cues of a scene, but some methods directly regressed control commands from a single image, and they did not utilize the geometric information of scenes~\cite{1,2,3,9}. In comparison, methods using depth maps are favorable~\cite{10,46,4,5} because controlling commands or path planning can be built upon this intermediate step, and the depth maps can be used for other tasks such as scene reconstruction. However, reliable depth estimation is computationally expensive, and the mentioned methods ran on larger platforms or off-board.

\subsection{Efficient Monocular Self-Supervised Depth Estimation}

In recent years single image depth estimation (SIDE) has attracted considerable attention from researchers with the development of deep learning. SIDE models using supervised training regress pixel-wise depth values from labeled data. As it requires additional effort to annotate the data, self-supervised depth estimation (SSDE) stands out and predicts depth from label-free monocular videos~\cite{18}. Subsequent work improved prediction accuracy by introducing multi-task learning~\cite{19,20}, adding uncertainty constraints~\cite{21,22}, or using more powerful deep learning architectures~\cite{23,24,25}. Some recent methods pursued the balance of model accuracy, speed, and size, which is also the focus of this paper. Fastdepth~\cite{26} adopted MobileNet~\cite{27} as the encoder, and achieved fast inference speed on embedded systems. However, this model was designed for supervised training. R-MSFM~\cite{28} designed a feature modulation module to learn multi-scale depth features, and reduced model size by controlling the encoder's layers. Lite-Mono~\cite{11} is a hybrid CNN and Transformer architecture, which is capable of extracting both enhanced local features and global contexts, and achieving state-of-the-art. It has a good trade-off between accuracy and model size ($3.1M$ parameters). Nonetheless, such a small model still exceeds the storage space of the GAP8 processor. To get a much smaller model the Lite-Mono network is streamlined and a model with $310K$ parameters is obtained in this paper. This results in the problem that the learning ability of this small model is limited. For the purpose of improving the learning ability of the model, the paper introduces knowledge distillation.

\subsection{Knowledge Distillation for Depth Estimation}

In a typical KD framework a larger~\textit{teacher} model transfers its knowledge to a smaller~\textit{student} model. The common way of KD can be through soft labels~\cite{12, 29} or intermediate feature-matching~\cite{30,31,32,33}. KD has been applied to the depth estimation task to boost lightweight models. Wang~\textit{et al.}~\cite{35} used the ResNet-101~\cite{37} as the teacher and MobileNet as the student, and set up the distillation between decoders of the two networks. Hu~\textit{et al.}~\cite{36} improved the knowledge distillation with auxiliary data. Pilzer~\textit{et al.}~\cite{34} also explored KD in depth estimation, but their method required stereo image pairs for training. However, the student models used in the above-mentioned research were still too heavy to be deployed on a GAP8. Some recent papers have pointed out that KD may have difficulty optimizing the student model and achieve unsatisfactory results, if there is a large learning capacity gap between teacher and student~\cite{38, 39}. Inspired by some KD methods for classification and semantic segmentation tasks~\cite{32,33,40} this paper proposes the CADiT module to encourage the student to pay attention to geometric cues from the teacher's feature channels, thus improving the KD process.

\section{METHOD}
\label{method}

\begin{figure}[!htb]
  \centering
\includegraphics[width=0.8\linewidth, height=3.5cm]{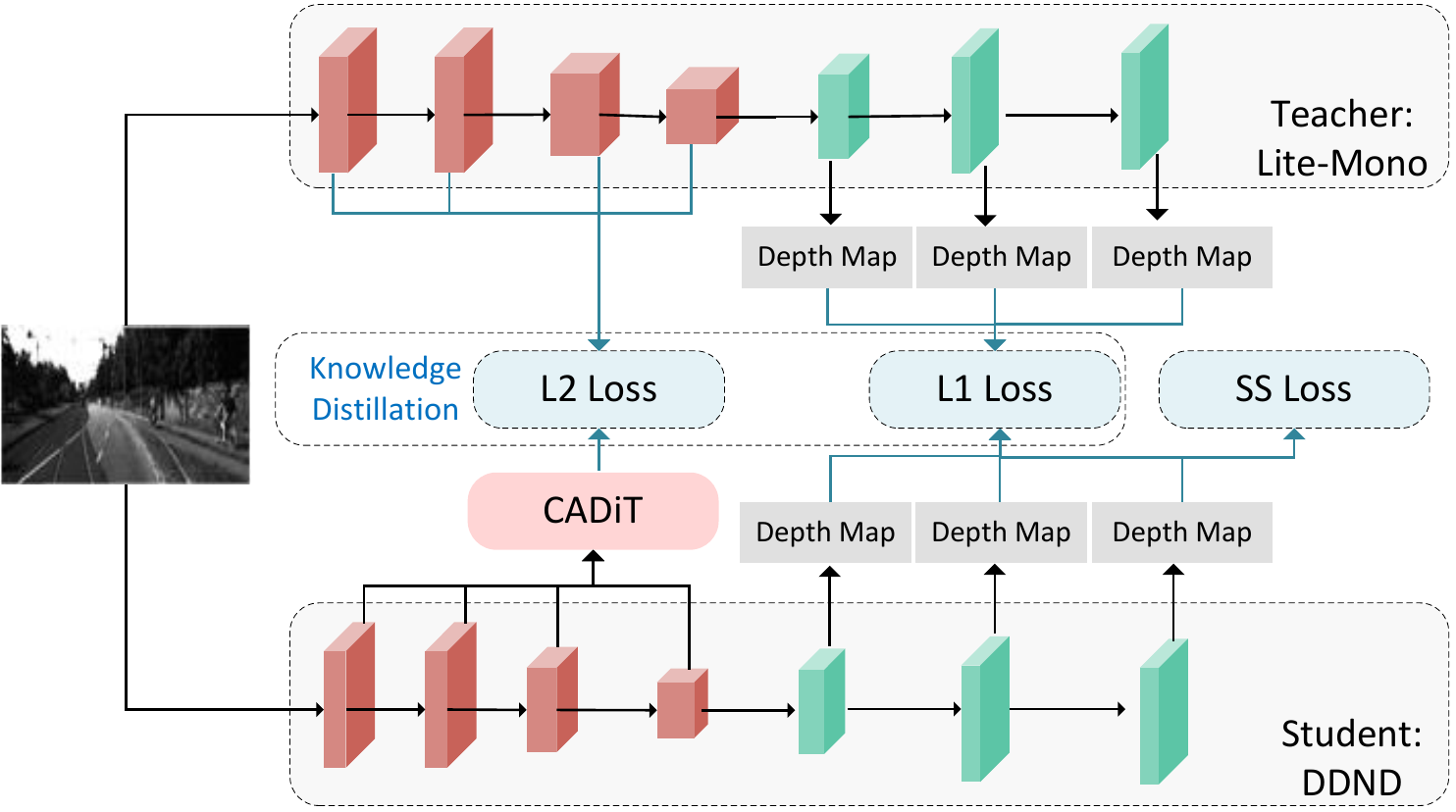}
   \caption{Overview of the proposed DDND. In addition to the self-supervised (SS) loss used in the SSDE training scheme, $L2$ loss and $L1$ loss are used to distill the teacher's knowledge into the student's encoder and decoder, respectively. The proposed CADiT is introduced in Section~\ref{sec:cadit}.}
   \label{fig:pipeline}
   \vspace{-5mm}
\end{figure}

The proposed Distilled Depth for Nano Drones (DDND) is shown in Fig.~\ref{fig:pipeline} and explained in detail in this section. First, the architecture of the network and the depth estimation training scheme are presented. Then, the KD scheme including the proposed CADiT is demonstrated. The last subsection introduces the control method using the generated depth map.

\subsection{Network Structures}
\begin{figure*}[!htb]
  \centering
  \vspace{0.2cm}
\includegraphics[width=0.6\linewidth]{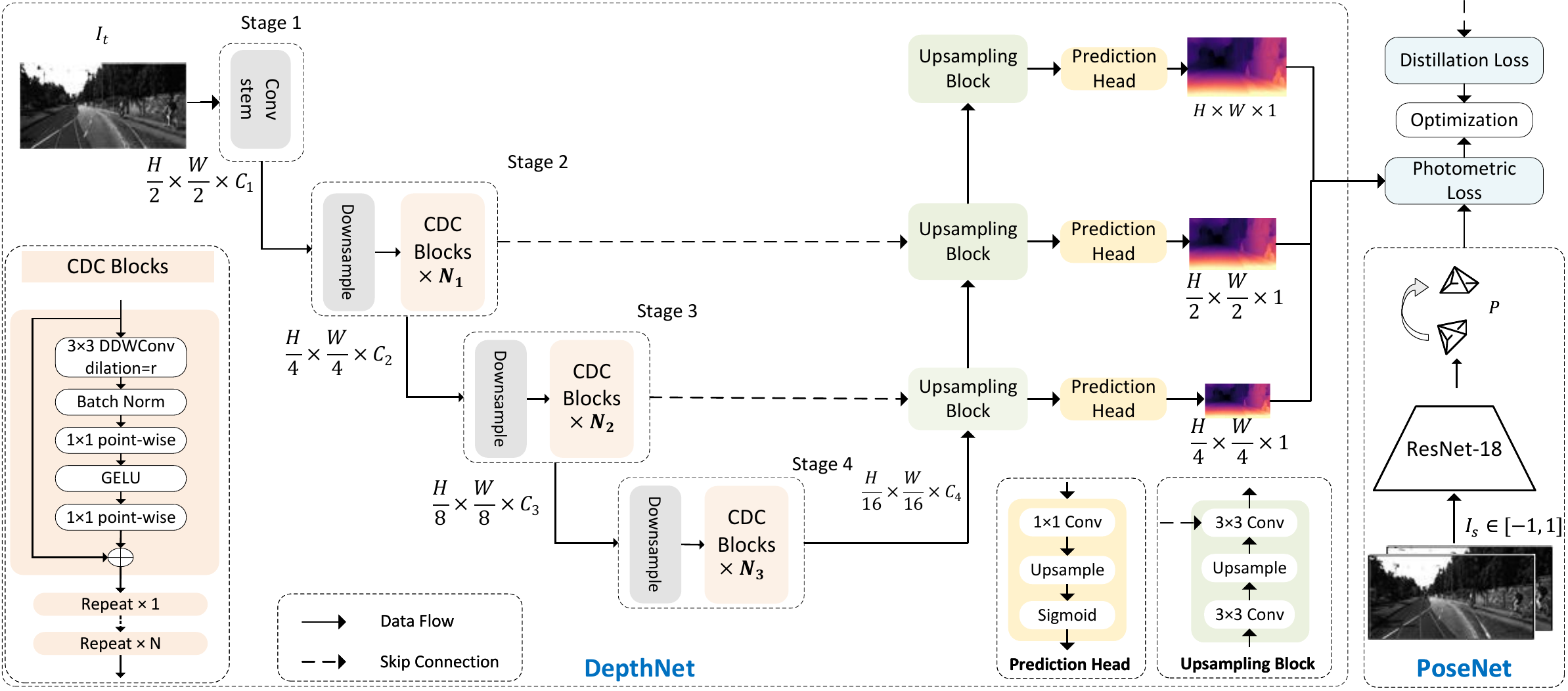}
   \caption{The training scheme of self-supervised depth estimation and the architectures of DepthNet and PoseNet.}
   \label{fig:network}
   \vspace{-6mm}
   
\end{figure*}

To make the model deployable on the GAP8 chip the encoder of Lite-Mono~\cite{11} is streamlined to reduce the number of trainable parameters. As shown in the upper part of Fig.~\ref{fig:network} the student model (DepthNet) is an encoder-decoder network, with four stages in the encoder to extract features. Its decoder concatenates features from the encoder and outputs inverse depth maps at three scales. The channel numbers of the encoder in Lite-Mono are [48, 48, 80, 128], while the student model used in this paper has channel numbers $[C1,C2,C3,C4]=[32, 32, 64, 80]$. 
The same dilation rates of Lite-Mono are used in the DDND network, and the total parameters are reduced from $3.1M$ to $310K$. The PoseNet is the same pre-trained ResNet-18 used in~\cite{18}, and it is not needed after training.

\subsection{SSDE Training Scheme}
Fig.~\ref{fig:network} also shows the self-supervised depth estimation (SSDE) training scheme, which aims at minimizing the photometric loss $\mathcal{L}_p$ between a target image $I_t$ and the reconstructed one $\hat{I_t}$. The DepthNet takes a gray-scale target image $I_t$ and predicts a depth map $D_t$. The PoseNet estimates the relative pose $P$ between the target image and an adjacent image $I_{t+s},  s\in[-1,1]$.

\subsubsection{Photometric Loss} With the camera's intrinsics $K$ known the photometric loss can be defined as:
\begin{equation}
 \mathcal{L}_p(\hat{I_t}, I_t) = \mathcal{L}_p(\mathcal{F}(I_{t+s}, P, D_t, K), I_t),
  \label{eq:lp}
\end{equation}
which can be calculated as a sum of SSIM (Structural Similarity Index) and the L1 loss between two images:
\begin{equation}
 \mathcal{L}_p(\hat{I_t}, I_t) = \alpha \frac{1-SSIM(\hat{I_t}, I_t)}{2}+(1-\alpha)\lVert \hat{I_t}-I_t\rVert,
  \label{eq:prl}
\end{equation}
with $\alpha$ being an empirical value of 0.85~\cite{18}. The minimum photometric loss and the auto-masking techniques~\cite{18} are adopted to improve the training. The final photometric loss is defined as:
\begin{equation}
\begin{aligned}
 \mathcal{L}_p(\hat{I_t}, I_t) =\left \langle\min\limits_{s\in [-1, 1]}\mathcal{L}_p(I_{t+s}, I_t) \textgreater \min\limits_{s\in [-1, 1]}\mathcal{L}_p(\hat{I_t}, I_t)]\right \rangle \\
 \cdot \min\limits_{s\in [-1, 1]}\mathcal{L}_p(\hat{I_t}, I_t),
  \label{eq:flp}
  \end{aligned}
  \vspace{-1mm}
\end{equation}
where $\langle\cdot \rangle$ operation outputs a binary mask to remove moving pixels.

\subsubsection{Smoothness Loss} The edge-aware smoothness loss~\cite{18} is also used:
\begin{equation}
 \mathcal{L}_{smooth} = \left|\partial_x\mathrm{d}_t^{\ast}\right|e^{-\left|\partial_xI_t\right|}+\left|\partial_x\mathrm{d}_t^{\ast}\right|e^{-\left|\partial_yI_t\right|},
  \label{eq:smooth}
\end{equation}
where $d_t^\ast = d_t/\hat{d_t}$ is the mean-normalized inverse depth. Therefore, the combined loss function for the self-supervised training is:
\begin{equation}
 \mathcal{L}_{ss} = \frac{1}{3}\sum_{j\in \{1, \frac{1}{2}, \frac{1}{4}\}}(\mathcal{L}_p+\lambda \mathcal{L}_{smooth}),
  \label{eq:tl}
\end{equation}
where $j$ can be three scales of the inverse depth. $\lambda$ is set to $1e^{-3}$ as in~\cite{18}.

\subsection{Knowledge Distillation Scheme}
\label{sec:cadit}
\subsubsection{Matching Intermediate Features using the Channel Aware Distillation Transformer (CADiT)}
Assume that the teacher network $T$ and the student network $S$ have intermediate feature maps denoted by $F_{T}\in \mathbb{F}^{H\times W\times C}$ and $F_{S}\in \mathbb{F}^{H\times W\times C^{'}}$, respectively. The feature map has a height of $H$, a width of $W$, and channel numbers of $C$. As shown in Fig.~\ref{fig:cadit} (a) the student's feature channels are increased to have the same channel numbers as the teacher. Then, they can be reshaped as $F_{T}\in \mathbb{F}^{N\times C}$ and  $F_{S}\in \mathbb{F}^{N\times C}$, respectively, where $N=H\times W$. The L2 loss can be used to minimize the discrepancy between the teacher's and student's intermediate features ($\mathcal{L}_{IF}$)~\cite{30}:
\begin{equation}
 \mathcal{L}_{IF} = ||F_S - F_T||_{2}.
  \label{eq:fm}
\end{equation}

However, such a direct feature-matching method may increase the difficulty of optimization if the student has poor learning ability. The proposed CADiT (Fig.~\ref{fig:cadit} (b)) allows each student channel to learn geometric cues from all the teacher's channels to improve feature-matching. Specifically, a $C\times C$ channel correlation map (CCM) is built between the transposed aligned student and the teacher:
\begin{figure}[!htb]
  \centering
\includegraphics[width=0.8\linewidth]{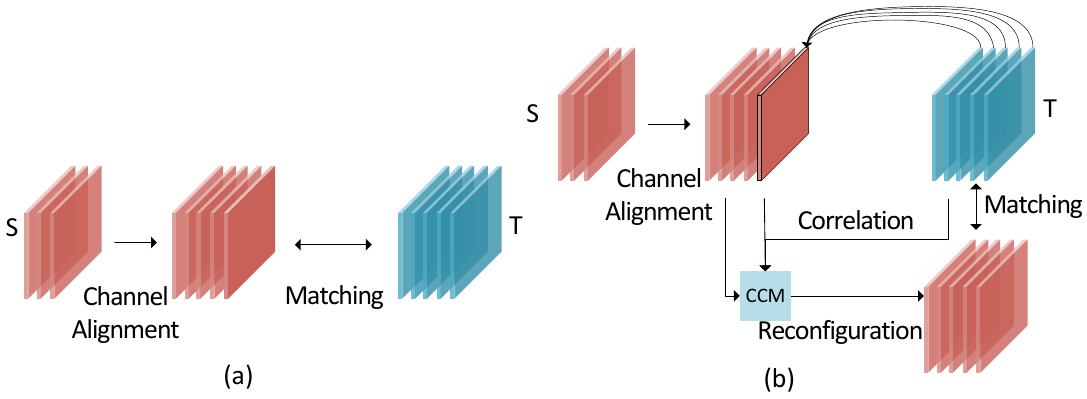}
   \caption{Intermediate feature-matching schemes. (a) is the conventional feature-matching scheme. (b) is the proposed CADiT that makes the student learn the channel correlations from the teacher.}
   \label{fig:cadit}
   \vspace{-6mm}
\end{figure}

\begin{equation}
CCM = Softmax(F_{S}^{\mathsf{T}} \cdot F_{T}) \in \mathbb{F}^{C\times C},
\end{equation}
where $(\cdot)$ is the inner product, and the CCM measures correlations between student and teacher channels. The student's features can be reconfigured as:
\begin{equation}
F_{S^{'}} = F_{S} + F_{S} \cdot CCM \in \mathbb{F}^{N\times C},
  \label{eq:s'}
\end{equation}
and the CADiT loss is computed as:

\begin{equation}
\mathcal{L}_{ CADiT} = ||F_{S^{'}} - F_T||_{2}.
  \label{eq:cadit}
\end{equation}

\subsubsection{Matching Outputs} As with the traditional KD methods~\cite{12,36,38} the L1 loss is also used to minimize the multi-scale depth maps generated by the teacher and the student. The $\mathcal{L}_{Out}$ is defined as:
\begin{equation}
\mathcal{L}_{Out} =\frac{1}{3}\sum_{j\in \{1, \frac{1}{2}, \frac{1}{4}\}} ||D_{S} - D_T||_{1}.
  \label{eq:out}
\end{equation}

The final loss function to train the network, combining Eq.~\ref{eq:tl}, is written as:
\begin{equation}
\mathcal{L} = \mathcal{L}_{ss} + \alpha\mathcal{L}_{CADiT} + \mathcal{L}_{Out},
  \label{eq:loss}
\end{equation}

where $\alpha$ is a weighting factor, set to 0.1 in this paper.

\subsection{Drone Controlling}
\begin{figure}[!htb]
  \centering
  \vspace{-3mm}
\includegraphics[width=0.7\linewidth]{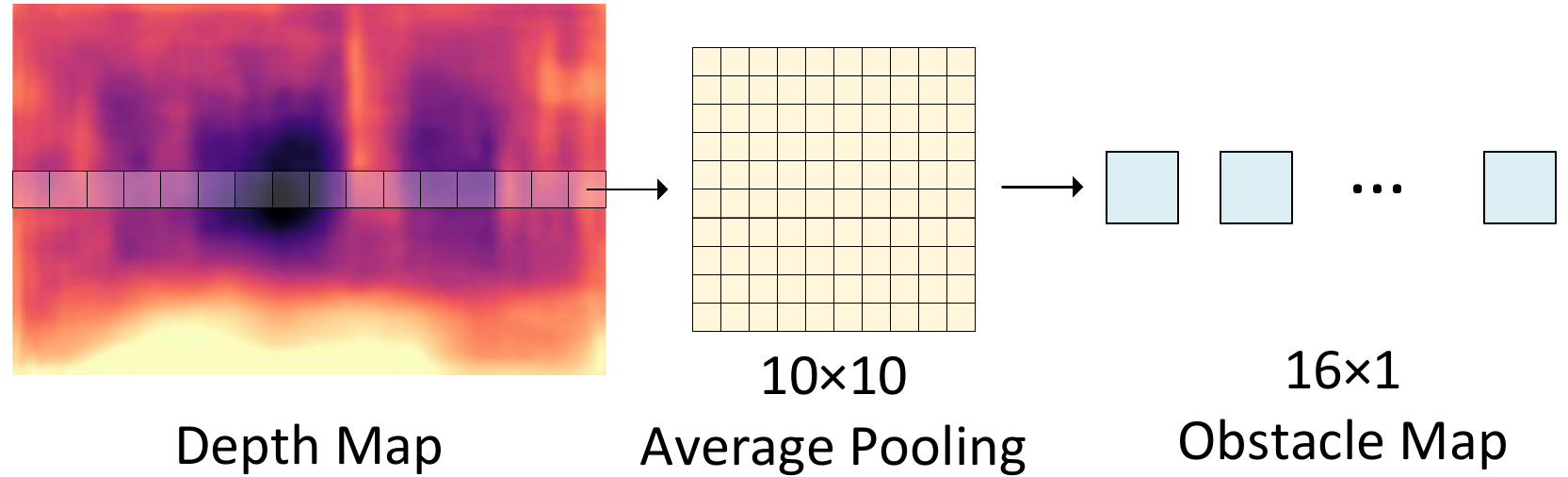}
   \caption{An obstacle map is generated by applying a sum-pooling operation on the horizontal center of the depth map. The depth values of all the pixels in each pooling window are averaged.}
   \label{fig:obstmap}
    \vspace{-5mm}
\end{figure}

With the depth map generated by the network the nano drone avoids obstacles based on the Behaviour Arbitration (BA) scheme~\cite{41}. Although this scheme was originally used in conjunction with sonas, it can also be used with depth maps. The behaviour $avoid$ is defined as:
\begin{equation}
\dot{\phi}=f_{avoid}(\phi)=\lambda_{avoid}\sum_{i}[(\phi-\psi_{i})\cdot e^{-c_{avoid}d_{i}}\cdot e^{-\frac{(\phi - \psi_{i})^2}{2\sigma_{i}^{2}}})],
  \label{eq:avoid}
\end{equation}
where $\lambda_{avoid}$ is a weighting factor, and $\phi$ is the current bearing. $d_{i}$ and $\psi_{i}$ are the depth value and direction of the $i-th$ value in the obstacle map, respectively. $\sigma_{i}$ is the field-of-view of the camera in the horizontal direction. The gain $c_{avoid}$ controls the sensitivity to obstacles. Lower gain allows the drone to react to obstacles further away. Increasing $\lambda_{avoid}$ changes the angular velocity of the drone more quickly. To construct the obstacle map the center rows of the depth map are \textit{average-pooled} with a window size of $10\times 10$ (Fig.~\ref{fig:obstmap}), resulting in a 1D obstacle map. The considerations of using such a simple control strategy are as follows: 1) The calculation is cheap, and no additional trainable parameters are required to generate the obstacle map; 2) It allows the selection of the next waypoint based on the depth map for path planning in future work. It is possible to use more complex control schemes, but this is beyond the scope of this paper.

\section{Experiments}
\label{sec:exp}
\subsection{Drone Platform}
\begin{figure}[!htb]
  \centering
  \vspace{-2mm}
\includegraphics[width=0.9\linewidth]{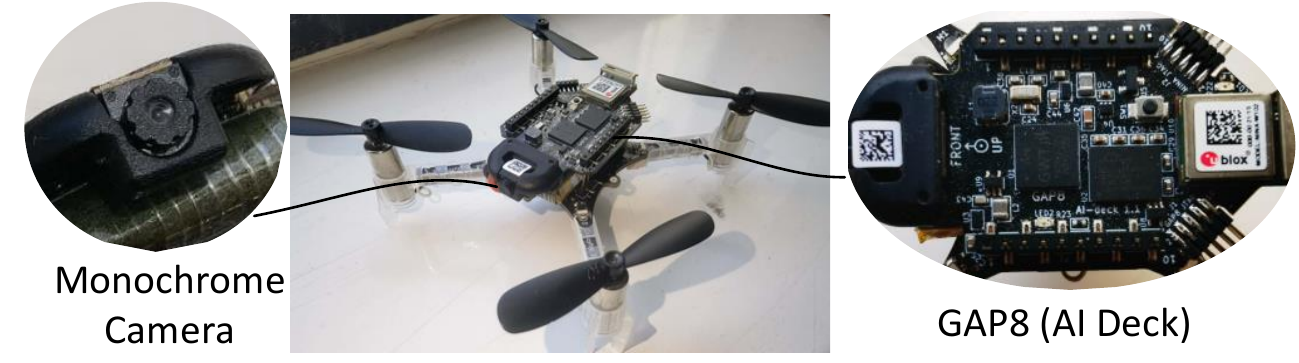}
   \caption{The Crazyflie is used as the drone platform in this paper.}
   \label{fig:drone}
   \vspace{-4mm}
\end{figure}

Bitcraze Crazyflie (Fig.~\ref{fig:drone}) is the drone platform used in this paper. It is equipped with a Flow Deck v2 at the bottom to measure the distance to the ground, and an ultra-low power GAP8 processor (AI Deck) integrating a monochrome camera (HM01B0-MNA). This drone platform weighs 34$g$. It has 22.65 GOPS of processing power and 512 KB of RAM.

\subsection{Datasets}
\label{sec:datasets}
\subsubsection{KITTI} KITTI is a multimodal dataset~\cite{42}, which consists of 61 stereo road scenes. In this paper the self-supervised model is trained on the Eigen split~\cite{43}. There are 39,180 monocular triplets used in the training, 4,421 for evaluation, and 697 for testing. During training all the RGB images in the KITTI dataset are resized to $192\times 640$ and converted to one-channel gray-scale images.

\subsubsection{Gray Campus Indoor} This in-house indoor drone dataset is collected by a Crazyflie in different buildings on our campus. It consists of 17 sequences, a total of 9,140 gray-scale images with an original resolution of $244\times 324$. Images are resized to $128\times160$ in this paper to meet the requirement of running speed.

\subsection{Implementation Details}
\label{sec:implementation}
\subsubsection{Network Training}
The proposed network is implemented in PyTorch. Models are trained for 30 epochs on KITTI, and 100 epochs on Gray Campus Indoor, with an NVIDIA TITAN Xp. The teacher model Lite-Mono is pre-trained on ImageNet~\cite{44} and then trained on KITTI. During training the teacher's weights are fixed, and only the student's weights are updated. AdamW~\cite{57} optimizer is used, and the weight decay is set to $1e^{-4}$. The initial learning rate is set to $1e^{-4}$ with a cosine learning rate schedule. 

\subsubsection{Model Quantization and Deployment} The trained PyTorch model is further converted to the ONNX (Open Neural Network Exchange) format and quantized using an 8-bit scheme, reducing the size of the weights from $747.6K$ bytes to $201.3K$ bytes. The controlling algorithm is implemented in C language. The Fabric Controller (FC) frequency of the GAP8 is set to $250 MhZ$.

\subsection{Results on Gray-Scale KITTI}
\begin{figure}[!htb]
  \centering
\includegraphics[width=1\linewidth]{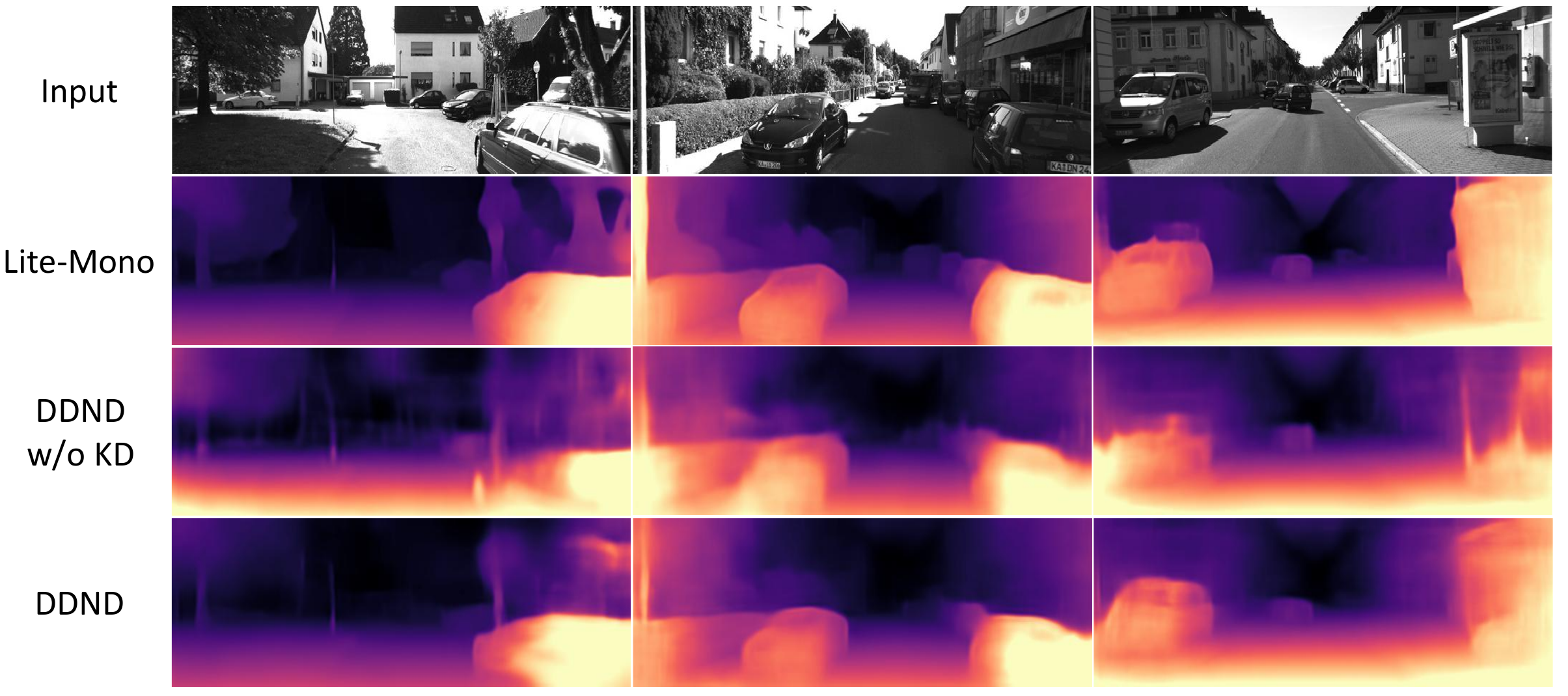}
   \caption{Qualitative results on KITTI. The student network can successfully learn feature representations from the teacher.}
   \label{fig:kittiresults}
   \vspace{-5mm}
\end{figure}

Table~\ref{exp:accuracy} shows the accuracy of models trained on the KITTI dataset, and the seven commonly used accuracy indicators~\cite{58} are AbsRel, SqRel, RMSE, RMSE log, $\delta <1.25$, $\delta <1.25^2$, and $\delta <1.25^3$. By comparing the results of Lite-Mono with Lite-Mono (RGB) it can be found that the self-supervised training based on the photometric loss also works on gray-scale images, albeit with less accuracy. This confirms the feasibility of the proposed method using gray-scale images for self-supervised depth estimation. Besides, results show that leveraging the proposed KD method in the training the accuracy is greatly improved. Fig.~\ref{fig:kittiresults} shows some images generated by the networks, and it can be observed that DDND learns knowledge from Lite-Mono and it is able to perceive larger objects. In addition, DDND can produce sharper depth maps at the edges of objects compared to the blurred depth maps produced without KD.

\begin{table}[ht!]
\begin{center}
\caption{Accuracy comparison on KITTI. "RGB" means that the training uses three-channel RGB images.}
\label{exp:accuracy}
\scalebox{0.58}{
\begin{tabular}{c|cccc|ccc|c}
\hline
\multirow{2}*{Method}&
\multicolumn{4}{c|}{Depth Error ($\downarrow$)}&
\multicolumn{3}{c|}{Depth Accuracy ($\uparrow$)}&\multirow{2}*{$\#$ Params.}\\

\cline{2-8}
& Abs Rel & Sq Rel & RMSE & RMSE log &$\delta <1.25$ &$\delta <1.25^2$&$\delta <1.25^3$\\
\cline{1-9}

Lite-Mono (RGB)~\cite{11}&0.107&0.765&4.561&0.183&0.886&0.963&0.983&3.1M\\
\cline{1-9}
Lite-Mono~\cite{11}&0.110&0.848&4.713&0.187&0.881&0.961&0.982&3.1M\\
DDND w/o KD & 0.157  &   1.259  &   5.593  &   0.229  &   0.796  &   0.930  &   0.973&0.31M  \\
DDND &   0.147  &   1.149  &   5.394  &   0.221  &   0.813  &   0.936  &   0.974&0.31M \\
\hline

\end{tabular}}
\end{center}
\end{table}

\subsection{Qualitative Results on Gray Campus Indoor}
\begin{figure}[!htb]
  \centering
\includegraphics[width=0.9\linewidth]{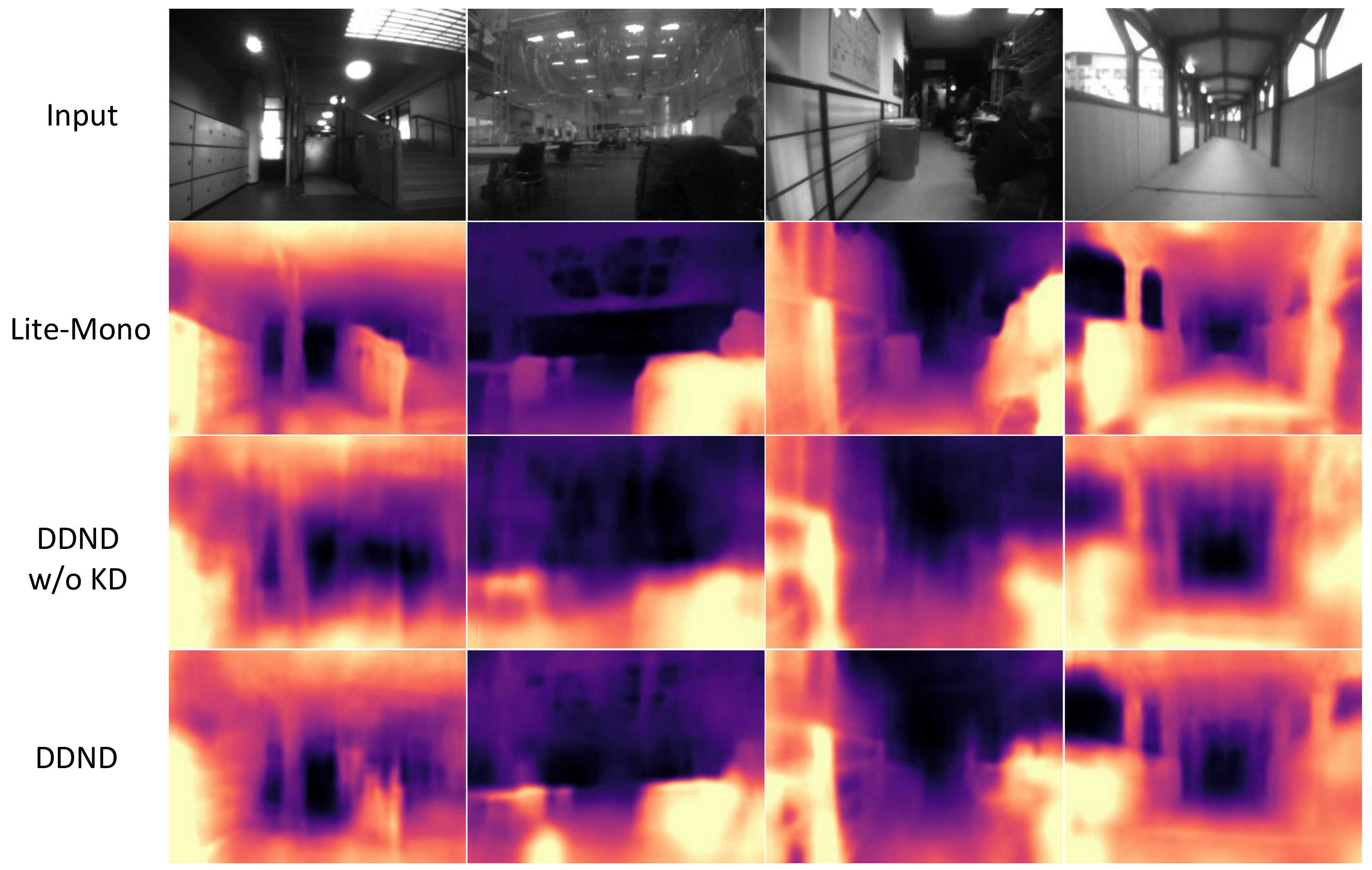}
   \caption{Qualitative results on Gray Campus Indoor.}
   \label{fig:gciresults}
   \vspace{-3mm}
\end{figure}

Fig.~\ref{fig:gciresults} shows some results on the in-house dataset. The dataset is challenging for the SSDE framework as it has many lighting sources and low-texture regions, such as walls and floors. In addition, scenes are more diverse. DDND benefits from the KD scheme and captures more detail in scenes.

\subsection{Ablation Study on KD Losses}
Ablation studies with different loss settings on KITTI are performed to validate the effectiveness of the proposed KD training and CADiT. Table~\ref{exp:ablation} shows that KD in the training improves the accuracy. Even if the proposed CADiT module is only used in the encoder, without the help of L1 loss in the decoder, good results can still be achieved (No.7).

\begin{table}[ht!]
\begin{center}
\caption{Accuracy comparison on KITTI of DDND using different KD losses. "E/D": KD in the encoder/decoder. "CD": channel loss proposed in~\cite{33}. The best results are highlighted in \textbf{bold}.}
\label{exp:ablation}

\scalebox{0.60}{
\begin{tabular}{c|c|cccc|ccc}
\hline
\multirow{2}*{No.}&
\multirow{2}*{KD Settings}&
\multicolumn{4}{c|}{Depth Error ($\downarrow$)}&
\multicolumn{3}{c}{Depth Accuracy ($\uparrow$)}\\

\cline{3-9}
&& Abs Rel & Sq Rel & RMSE & RMSE log &$\delta <1.25$ &$\delta <1.25^2$&$\delta <1.25^3$\\
\cline{2-9}

1&E: n/a, D: n/a & 0.157  &   1.259  &   5.593  &   0.229  &   0.796  &   0.930  &   0.973  \\
\cline{2-9}

2&E: n/a, D: L1&0.155  &   1.186  &   5.502  &   0.231  &   0.790  &   0.929  &   0.973  \\

3&E: L2, D: L1 &   0.154  &   1.305  &   5.653  &   0.229  &   0.803  &   0.932  &   0.972  \\

4&E: CD, D: n/a & 0.155  &   1.242  &   5.649  &   0.228  &   0.797  &   0.930  &   0.973  \\

5&E: CD, D: L1&0.152  &   1.208  &   5.561  &   0.226  &   0.807  &   0.934  &   0.973  \\

6&E: n/a, D: CADiT+L1&0.149  &   1.172  &   5.472  &   0.226  &   0.807  &   0.933  &   \textbf{0.974}  \\

7&E: CADiT, D: n/a&0.149  &   1.236  &   5.528  &   0.223  &   \textbf{0.815}  &   \textbf{0.936}  &   0.973  \\

\cline{2-9}
\textbf{8}&E: CADiT, D: L1 &   \textbf{0.147}  &   \textbf{1.149}  &   \textbf{5.394}  &   \textbf{0.221}  &   0.813  &   \textbf{0.936}  &   \textbf{0.974} \\
\hline

\end{tabular}}
\end{center}
\end{table}
\vspace{-4mm}

\subsection{Test in Real Environments}
\begin{figure}[thpb]
      \centering
      \vspace{0.2cm}
      \includegraphics[width=0.9\linewidth]{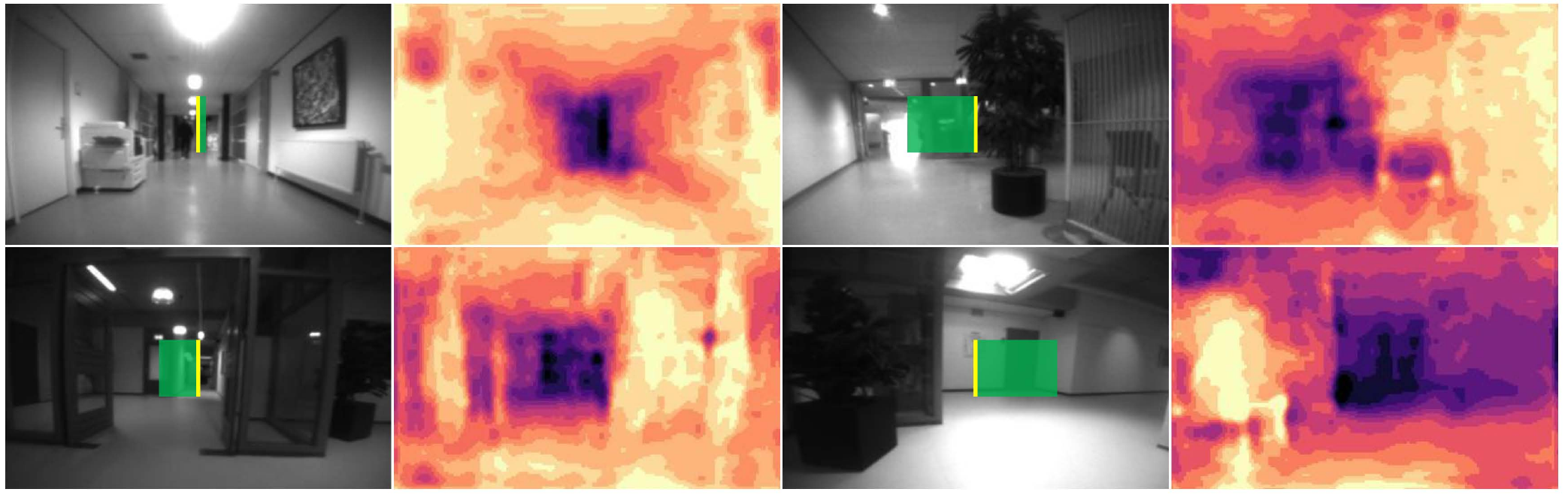}
      
      \caption{Real environment tests. Gray-scale input images and their corresponding depth maps generated by the quantized CNN model are shown. The green bar in each gray-scale image denotes the change in angular velocity to avoid obstacles.}
      \label{fig:real}
      \vspace{-3mm}
   \end{figure}

The proposed approach is tested in real indoor environments. Fig~\ref{fig:real} shows some images taken by the nano drone, and the generated depth maps with the deployed quantized model. The green bars on the gray-scale images denote steering commands for avoiding obstacles calculated by Eq.~\ref{eq:avoid}. Due to model quantization, the on-board network is not able to generate smooth depth maps, but these maps still succeed in showing the structures and obstacles of these scenes. Considering the inference speed of GAP8 the $c_{avoid}$ defined in Eq.~\ref{eq:avoid} is set to 0.1 to make sure that the drone is able to react to obstacles at a safer distance.

\subsection{Failure Cases}
\begin{figure}[thpb]
      \centering
      \includegraphics[width=0.9\linewidth]{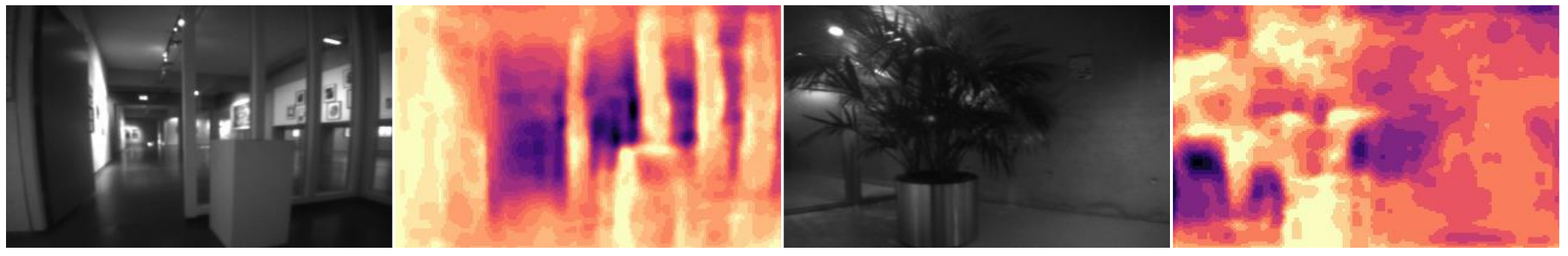}
      \caption{The method fails to avoid obstacles in areas of glass and when closed to walls.}
      \label{fig:failure}
      \vspace{-5mm}
   \end{figure}

The proposed method fails to estimate the depth of the glass or if it is too close to a wall, as shown in Fig.~\ref{fig:failure}. This is also a limitation of SSDE methods, and this problem can be overcome by integrating additional sensors, such as ultrasonic sonar, to detect the distance to glass and walls.

\subsection{Inference Speed Analysis}
The inference speed of the proposed network is evaluated both on the NVIDIA TITAN Xp GPU (Graphics Processing Unit) and the GAP8 processor. As shown in Table~\ref{exp:speed} there is little difference in the speed of the network inferring on TITAN Xp at either resolution, but on GAP8 it is about 6 times faster for the resolution of $128\times160$. It can also be observed that the computing power of edge computing devices such as GAP8 is extremely limited. The inference speed of 1.24 FPS is acceptable because the nano drone flies at a low speed during tests.

\begin{table}[ht!]
\begin{center}
\caption{Inference speed evaluation under two image resolutions.}
\label{exp:speed}
\scalebox{0.8}{
\begin{tabular}{c|cc}
\hline
\multirow{2}*{Resolution}&
\multicolumn{2}{c}{Speed (FPS)}\\

&NVIDIA TITAN Xp & GAP8 \\
\cline{1-3}

$128\times 160$&434.78&1.24\\

$224\times 320$&431.53&0.22\\

\hline

\end{tabular}}
\end{center}
\end{table}
\vspace{-5mm}

\section{CONCLUSIONS}
\label{sec:conclusion}
This paper proposes a lightweight depth estimation framework DDND, for obstacle avoidance on nano drones. To enhance the learning ability of this small network this paper integrates knowledge distillation and proposes the CADiT module for better knowledge transfer. The quantitative and qualitative results on the KITTI dataset validated the effectiveness of the proposed KD module. The model is then quantized so that it can infer on a Crazyflie for real environment tests. Further work will be focused on selecting waypoints on the generated depth maps for path planning and integrating off-board 3D construction for better unknown space exploration.












\bibliographystyle{IEEEtran}
\bibliography{IEEEabrv}

\end{document}